\pdfoutput=1

\documentclass[11pt,a4paper]{article}
\usepackage[hyperref]{acl2021}
\usepackage{times}
\usepackage{latexsym}
\usepackage{multicol}
\usepackage{graphicx}
\usepackage{booktabs}
\usepackage{tabularx}
\usepackage[utf8x]{inputenc} 
\usepackage{xspace}

\newcommand{\GT}[0]{{gT5}\xspace}
\newcommand{\lang}[0]{{\scshape Lang-8}\xspace}
\newcommand{\clangg}[0]{{\scshape cLang-8}\xspace}
\newcommand{\clangs}[0]{{\scshape cLang-8-s}\xspace}
\newcommand{\beatrain}[0]{BEA\xspace}
\newcommand{\INSERTION}[0]{{\scshape Felix}\xspace} 
\newcommand{\fzero}[0]{F\textsubscript{0.5}\xspace}
\newcommand{\conll}[0]{CoNLL-14\xspace}
\newcommand{\beatest}[0]{BEA test\xspace}

\newcommand*\rot{\rotatebox{60}}

\usepackage{microtype}

\aclfinalcopy 

\title{A Simple Recipe for Multilingual Grammatical Error Correction}
\author{Sascha Rothe \\
  Google \\
  \texttt{rothe@google.com} \\\And
  Jonathan Mallinson \\
  Google \\
  \texttt{jonmall@google.com} \\\And
  Eric Malmi \\
  Google \\
  \texttt{emalmi@google.com} \\\AND
  Sebastian Krause \\
  Google \\
  \texttt{bastik@google.com} \\\And
  Aliaksei Severyn \\
  Google \\
  \texttt{severyn@google.com} \\}

\date{}

\begin{document}
\maketitle
\begin{abstract}
This paper presents a simple recipe to train state-of-the-art multilingual Grammatical Error Correction (GEC) models. We achieve this by first proposing a language-agnostic method to generate a large number of synthetic examples. The second ingredient is to use large-scale multilingual language models (up to 11B parameters). Once fine-tuned on language-specific supervised sets we surpass the previous state-of-the-art results on  GEC benchmarks in four languages: English, Czech, German and Russian. Having established a new set of baselines for GEC, we make our results easily reproducible and accessible by releasing a \clangg dataset.\footnote{\clangg can be found at \url{https://github.com/google-research-datasets/clang8}} It is produced by using our best model, which we call \GT, to clean the targets of a widely used yet noisy \lang dataset. \clangg greatly simplifies typical  GEC training pipelines composed of multiple fine-tuning stages -- we demonstrate that performing a single fine-tuning step on \clangg with the off-the-shelf language models  yields further accuracy improvements over an already top-performing \GT model for English. 
\end{abstract}

\section{Introduction}

Grammatical Error Correction (GEC) is the task of correcting grammatical and other related errors in text.
It has been the subject of several modeling efforts in recent years due to its ability to improve grammaticality and readability of user generated texts.
This is of particular importance to non-native speakers, children, and individuals with language impairments, who may be more prone to producing texts with grammatical errors.

Modern approaches often view the GEC task as monolingual text-to-text rewriting \cite{naplava,katsumata2020stronger,grundkiewicz-etal-2019-neural} and employ encoder-decoder neural architectures~\cite{sutskever2014sequence,bahdanau2014neural}.
These methods typically require large training sets to work well~\cite{malmi-etal-2019-encode} which are scarce especially for languages other than English.
One of the largest and most widely used datasets for GEC is the \lang Learner Corpus, which covers 80 languages and has been created by language learners correcting each other's texts.\footnote{Corpus collected from \url{https://lang-8.com/}}
However, the distribution of languages is very skewed, with Japanese and English being the most prevalent languages with over a million ungrammatical-grammatical sentence pairs each, while only ten languages have more than 10,000 sentence pairs each.
Additionally, given the uncontrolled nature of the data collection, many of the examples contain unnecessary paraphrasing and erroneous or incomplete corrections. 

Limited amounts of suitable training data has led to multiple approaches that propose to generate synthetic training data for GEC~\cite{madnani2012exploring,grundkiewicz2014wiked,grundkiewicz-etal-2019-neural,lichtarge-etal-2019-corpora,awasthi-etal-2019-parallel}.
Although using synthetic data as the first fine-tuning step has been shown to improve model accuracy, it introduces practical challenges that make the development and fair comparison of GEC models challenging: ($i$) the synthetic methods often require language-specific tuning (e.g. language-specific hyperparameters and spelling dictionaries~\cite{naplava}), and; ($ii$) due to the inability of synthetic data to capture the complete error distribution of the target eval sets, the final model is obtained by following a multi-stage fine-tuning process~\cite{lichtarge-etal-2019-corpora,lichtarge-etal-2020-data,omelianchuk-etal-2020-gector}. 
Because of this, carefully picking the learning rates and number of training steps for each of the fine-tuning stages is required, making it difficult to replicate and build on top of previous best reported models.

The ideas of leveraging self-supervised pre-training and increasing the model size have yielded significant improvements on numerous seq2seq tasks in recent years~\cite{t5,mt5,BART,MASS,KERMIT,Bert2Bert}, but these approaches have been applied to GEC to only a limited extent.

In this paper we adopt the mT5 \citep{mt5} as our base model which has already been pre-trained on a corpus covering 101 languages.
To adapt the model to the GEC task, we design a fully unsupervised language-agnostic pre-training objective that mimics corrections typically contained in labeled data.
We generate synthetic training data by automatically corrupting grammatical sentences, but in contrast to the previous state-of-the-art by \citet{naplava} for low-resources languages, we use our synthetic pre-training to train a single model on all 101 languages, employing no language-specific priors to remain fully language-agnostic.  After pre-training we further fine-tune our model on supervised GEC data for available languages (with data conditions ranging from millions to tens of thousands).
Additionally, we explore the effect of scaling up the model size from 60M to 11B parameters. We surpass the previous state-of-the-art results on four evaluated languages: English, Czech, German and Russian.

Fine-tuning and running inference with our largest and most accurate models require multi-GPU/TPU infrastructure. To make the results of our research widely accessible we release a \clangg dataset obtained by using our largest \GT model to clean up the targets of the frequently used yet noisy \lang dataset. We show that off-the-shelf variants of T5~\cite{t5} when fine-tuned only on \clangg, outperform those models trained on the original \lang data with and w/o additional fine-tuning data, thus simplifying the complex multi-stage process of training GEC models. Thus \clangg not only allows others to easily train highly competitive GEC models, but it also greatly simplifies GEC training pipeline, basically reducing a multi-step fine-tuning processs to a single fine-tuning step.

Our contributions in this paper are three-fold: (1)~We show that a simple language-agnostic pre-training objective can achieve state-of-the-art GEC results when models are scaled up in size; (2)~We show the effect model size has on GEC, and; (3)~We release a large multilingual GEC dataset based on Lang-8, which allows for state-of-the-art results without additional fine-tuning steps, thus significantly simplifying the training setup.

\section{Model}
Our model builds on top of mT5~\citep{mt5} a multilingual version of T5 \citep{t5} -- a Transformer encoder-decoder model which has been shown to achieve state-of-the-art results on a wide range of NLG tasks. 
mT5 comes in different sizes, however for this work we use \emph{base} (600M parameters) and \emph{xxl} (13B parameters).

\subsection{mT5 Pre-training}\label{task-agnostic-pre-training}
mT5 has been pre-trained on mC4 corpus, a subset of Common Crawl, covering 101 languages and composed of about 50 billion documents. 
For details on mC4, we refer the reader to the original paper \citep{mt5}.
The pre-training objective is based on a span-prediction task, an adaptation of masked-language objective for autoregressive seq2seq models.
An example of span prediction:\\
\\
{\footnotesize
\setlength{\parindent}{3pt}
\begin{tabularx}{\columnwidth}{lX}
\textbf{Input:} & \texttt{A Simple} [x] \texttt{Multilingual Grammatical Error} [y] \\[1.3em]
\textbf{Target:} & [x] \texttt{Recipe for} [y] \texttt{Correction} \\[1em]
\end{tabularx}
}

{\setlength{\parindent}{0pt}
 All mT5 models were trained for 1M steps on batches of 1024 input sequences with a maximum sequence length of 1024, corresponding to roughly 1T seen tokens.}
For all of our experiments we use the publicly available mT5 and T5 checkpoints (Section \ref{lang8} only).

\subsection{GEC Pre-training}\label{gec-pre-training}
The span-prediction objective of mT5 does not enable the model to perform GEC without further fine-tuning, as the span-prediction task uses  special  tokens  to  indicate  where  text should be inserted.
Another limiting constraint is that  mT5 has  been trained on paragraphs, not sentences.
We therefore split all paragraphs in mC4 corpus into sentences.
We corrupt each sentence using a combination of the following operations:
a) drop spans of tokens b) swap tokens c) drop spans of characters d) swap characters e) insert characters\footnote{We insert characters from the same passage, thus avoiding to insert character from a different alphabet.} f) lower-case a word g) upper-case the first character of a word.
An example pair of an original sentence and its corrupted version looks as follows: \\
\\
{\footnotesize
\setlength{\parindent}{3pt}
\begin{tabularx}{\columnwidth}{lX}
\textbf{Input:} &
\texttt{Simple recipe for Multingual Grammatical Correction Error} \\[1.3em]
\textbf{Target:} &
\texttt{A Simple Recipe for Multilingual Grammatical Error Correction}
\\[1em]
\end{tabularx}
}

{\setlength{\parindent}{0pt}
We leave about 2\% of examples uncorrupted, so the model learns that inputs can also be grammatical.}
We refrain from using more sophisticated text corruption methods, as these methods would be hard to apply to all 101 languages.
For example, \citet{naplava} perform word substitutions with the entries from ASpell\footnote{\url{http://aspell.net}} which in turn makes the generation of synthetic data language-specific.
Pre-training with this unsupervised objective is done on all languages in the mC4 corpus and not limited to the languages evaluated in this paper.

\section{\GT: Large Multilingual GEC Model} 
\paragraph*{Fine-tuning datasets.}\label{fine-tuning-data}
For English, we fine-tune our pre-trained models on the FCE \citep{fcedataset} and W\&I \citep{bea-2019} corpora.
For Czech, German, and Russian, we use the AKCES-GEC \citep{naplava}, Falko-MERLIN \citep{boyd-2018}, and RULEC-GEC \citep{rozovskaya-roth-2019-grammar} datasets, respectively. Table~\ref{tbl:dataset} reports statistics of datasets available for different languages.

\begin{table}[h]
\footnotesize
\begin{tabularx}{\columnwidth}{lXlll}
\toprule 
lang & Corpus & Train & Dev & Test \\ \midrule 
EN   & FCE, W\&I    & 59,941 &  &  \\
EN   & CoNLL-13/-14 &        & 1,379 & 1,312 \\
EN   & BEA          &        &       & 4,477 \\
CS   & AKCES-GEC    & 42,210 & 2,485 & 2,676       \\
DE   & Falko-MERLIN & 19,237 & 2,503 & 2,337       \\
RU   & RULEC-GEC    &  4,980 & 2,500 & 5,000       \\
\bottomrule 
\end{tabularx}
\caption{The size of the datasets used to fine-tune \GT.}
\label{tbl:dataset}
\end{table}

\paragraph*{Training Regime.}
We experimented with several training setups. All of them build on the mT5 pre-trained models (Section \ref{task-agnostic-pre-training}). We experimented with
a) mixing GEC pre-training data (Section \ref{gec-pre-training}) with fine-tuning data (Section \ref{fine-tuning-data}),
b) mixing pre-training and finetuning examples but annotating them with different prefixes, and
c) first using GEC pre-training until convergence and then fine-tuning.
While c) is the most computationally expensive approach, it also gave us the best results.
GEC pre-training as well as finetuning uses a constant learning rate of $0.001$.
Pre-training is done until convergence and fine-tuning until exact match accuracy on the development set degrades, which happens after 200 steps or 800k seen examples or 7 epochs.

\paragraph*{Results.}
For English, we evaluate on standard benchmarks from \conll (noalt)\footnote{For \conll, we fix tokenization discrepancies by post-processing model outputs with a set of heuristics. \textbf{UPDATE(2022-07-25):} We have uploaded a simplified version of the post-processing heuristics at \url{https://github.com/google-research-datasets/clang8/blob/main/retokenize.py} and updated the T5 results in Table~\ref{tbl:clang8-results} to use the simplified heuristics.} and the \beatest \cite{bea-2019}, while we use CoNLL-13 as the development set (Table \ref{tbl:dataset}).
For other languages we use the test and development sets associated with their training data.
Table~\ref{tbl:results} shows the results for all languages. 
We first see that the base model size is inferior to the current state-of-the-art models.
This is expected as the model capacity is not enough to cover all 101 languages.
We therefore use a larger xxl (11B) model, which produces new state-of-the-art results on all languages except for English.
When looking at the development set performance for English, we observed that it had a high variance and the training was over-fitting very quickly.
This suggests that train and dev/test set domains are not well aligned for English.
In the following Section \ref{lang8} we further refine our approach, also achieving state-of-the-art results for English.

\begin{table}[t]
\centering
\resizebox{\columnwidth}{!}{%
\begin{tabular}{llllll}
\toprule 
Models    & \rot{\conll}  & \rot{\beatest} & \rot{Czech} & \rot{German} & \rot{Russian}   \\ \midrule
 \citeauthor{omelianchuk-etal-2020-gector}$^*$  & 66.5 & \textbf{73.6} & -  & - & - \\
 \citeauthor{lichtarge-etal-2020-data}$^*$ & \textbf{66.8} & {73.0} & - & - & - \\
 \citeauthor{naplava}    & 63.40 & 69.00 & 80.17 & 73.71 & 50.20 \\
 \citeauthor{katsumata2020stronger}$^*$ & 63.00 & 66.10 & 73.52 &  68.86&   44.36 \\
 \midrule 
 \GT base   & 54.10 & 60.2	  & 71.88 & 69.21 & 26.24 \\ 
 \GT xxl     & 65.65 & 69.83	  & \textbf{83.15} & \textbf{75.96} & \textbf{51.62} \\
 \bottomrule 
\end{tabular}
}
\caption{\fzero Scores. 
Models denoted with {}$^*$ are ensemble models. We used the $M^2$ scorer for \conll, Russian, Czech and German, and the ERRANT scorer \cite{bryant-etal-2019-bea} for \beatest.}
\label{tbl:results}
\end{table}

\section{\clangg: Cleaned \lang Corpus} \label{lang8}
To be able to distill the knowledge learned by gT5 xxl into smaller, more practical models, we create and release \clangg, a cleaned version of the popular \lang corpus.
As discussed earlier, \lang is a large corpus of texts written by language learners and user-annotated corrections to these texts. 
However, corrected texts frequently contain unnecessary paraphrasing and erroneous or incomplete corrections -- phenomena that hurt the performance of a GEC model trained on this data.
For instance, the following source--target pair is taken from \lang: \textit{``It is cloudy or rainy recently~.''} $\rightarrow$ \textit{``It is It 's been cloudy or and rainy recently~.''}

We experiment with two approaches for cleaning the data. First, to create \clangg, we generate new targets for \lang, disregarding the original targets.
We tried using both the unsupervised model, which was trained using the GEC pre-training objective (Section \ref{gec-pre-training}) and the supervised model (\GT xxl) (Section \ref{fine-tuning-data}), but the former did not yield comparable results, so all reported numbers use the supervised model.
Second, to create \clangs, we used the unsupervised and the supervised models to score the original targets, disregarding the lowest scoring 20\%, 50\%, 70\%, or 90\% targets.
Disregarding 50\% was the best performing setup and there was not a significant difference between the supervised and unsupervised model.
We therefore report numbers using the unsupervised model disregarding the worst 50\% of the targets.
Table \ref{tbl:lang-8-stats} shows that \clangg moderately reduces the Word Error Rate (WER) between the source and target, with deletions receiving the largest relative reduction, which may suggest that less information from the source sentence is removed.
In contrast \clangs has a significantly lower WER, indicating that the unsupervised model has only kept corrections which are close to the source sentence.  

\begin{table}[t]
\footnotesize
\begin{tabularx}{\columnwidth}{Xccccc}
\toprule 
       & LR & WER & Sub  &  Del  &  Ins    \\ \midrule 
\lang &  98\%    &   15.46  &  8.85 & 2.41 & 4.19      \\
 \clangg      & 98\% &  10.11 &  5.85 & 1.35 & 2.92       \\ 
 \clangs & 99\% &01.22 & 0.64& 0.00 & 0.58 \\
 \bottomrule 
\end{tabularx}
\caption{Dataset statistics  of English \lang and \clangg, including sequence \textbf{L}ength \textbf{R}atio between the source and the target, \textbf{W}ord \textbf{E}rror \textbf{R}ate, which is comprised of \textbf{Sub}stitutions, \textbf{Del}etions, and \textbf{Ins}ertions.  }
\label{tbl:lang-8-stats}
\end{table}

\paragraph*{Experiments.}
To evaluate the effect cleaning \lang has for English, we train two distinct models on this data: T5~\cite{t5}, a monolingual sequence-to-sequence model, and \INSERTION~\cite{mallinson-etal-2020-felix}, a non-auto-regressive text-editing model.\footnote{The \textsc{FelixInsert} variant which we use does not employ re-ordering.}
We also tried fine-tuning these models on \beatrain{} (i.e. FCE and W\&I) after fine-tuning them on \clangg, but this did not further improve the scores but slightly decreased them, e.g. $0.43$ absolute decrease for BEA test when using T5 base.
This can be explained by the fact that the model used to clean the target texts has already been trained on \beatrain.
This suggests that the typical GEC training pipeline where a model is first fine-tuned on \lang and then on \beatrain{} can be both simplified and made more accurate by only fine-tuning on \clangg.

Finally, we train mT5 models on the German and Russian portions of the \clangg dataset and evaluate these models on the test sets from Table~\ref{tbl:dataset}.

\begin{table}[t]
\footnotesize
\setlength{\tabcolsep}{5pt}
\begin{tabularx}{\columnwidth}{llXll}
\toprule
Model & \#params & Training Data & \rot{\conll}  & \rot{\beatest}  \\
\midrule
SOTA  &  & & 66.8 & 73.6  \\ 
gT5 xxl & & & 65.65 & 69.83 \\
\midrule
\INSERTION  & 220M & \lang & 41.63 &  30.54 \\
\INSERTION  & 220M & \lang + \textsc{BEA} & 48.75 & 48.80  \\ 
\INSERTION  & 220M & \clangg & \textbf{58.21} & \textbf{59.05} \\ 
\midrule
T5 base & 220M & \lang & 50.52 & 59.14	 \\
T5 base & 220M & \lang + \textsc{BEA}  & 60.56 & 67.12	 \\
T5 base & 220M & \clangg & \textbf{65.05} & \textbf{69.38} \\ 
T5 base & 220M & \clangs & 58.70 & 59.95 \\ 
\midrule
T5 small & 60M & \clangg & 60.54 & 65.01 \\ 
T5 base  & 220M & \clangg & 65.05 &  69.38	 \\ 
T5 large & 770M & \clangg & 66.04 & 72.06 \\ 
T5 xl    & 3B & \clangg & 67.65 & 73.92 \\ 
T5 xxl   & 11B & \clangg & \textbf{68.75} & \textbf{75.88} \\ 
\bottomrule
\end{tabularx}
\caption{\fzero scores on \conll and \beatest. Block two and three compare different training data. The last block compares different model sizes for T5.}
\label{tbl:clang8-results}
\end{table}

\paragraph*{Results \& Analysis.}
The results for \conll and \beatest benchmarks can be seen in Table~\ref{tbl:clang8-results}.
For both models and both test datasets, \clangg improves the \fzero score compared to using the original \lang corpus.
While \clangs performs significantly worse than \clangg, it still improves over \lang.
In terms of model size, larger models are consistently better then their smaller siblings.
This is even true when comparing xl and xxl, suggesting that there might still be headroom by using models larger than xxl.

\begin{table}[t]
\footnotesize
\begin{tabularx}{\columnwidth}{XXXXX}
\toprule
 & \multicolumn{2}{c}{\lang} & \multicolumn{2}{c}{\clangg} \\
 & base & xxl & base & xxl \\ \midrule 
PUNCT &  68.27 & \textbf{78.75} & 75.51 & {76.31} \\
DET    & 63.84 & 77.31 & 79.04 & \textbf{83.88} \\
PREP&    57.09 & 72.54 & 74.67& \textbf{79.79} \\
ORTH &   72.77 & \textbf{76.86} & 69.23& 71.39 \\
SPELL &  74.38 & 84.64 & {85.83} & \textbf{88.29} \\
\bottomrule 
\end{tabularx}
\caption{\beatest scores for the top five error types.  Bold scores represent the best score for each error type.}
\label{tbl:error}
\end{table}

\begin{table}[t]
\footnotesize
\setlength{\tabcolsep}{5pt}
\begin{tabularx}{\columnwidth}{llXll}
\toprule
Model & \#params & Training Data & \rot{German}  & \rot{Russian}  \\
\midrule
SOTA  &  & & 73.71 & 50.20  \\ 
gT5 xxl & & & \textbf{75.96} & \textbf{51.62} \\
\midrule
mT5 small & 300M & \clangg & 61.78 & 17.80 \\ 
mT5 base  & 580M & \clangg & 67.19 & 25.20 \\ 
mT5 large & 1.2B & \clangg & 70.14 & 27.55 \\ 
mT5 xl    & 3.7B & \clangg & 72.59 & 39.44 \\ 
mT5 xxl   & 13B & \clangg & 74.83 & 43.52 \\  
\bottomrule
\end{tabularx}
\caption{\fzero scores on German and Russian.}
\label{tbl:clang8-mt5-results}
\end{table}
In Table \ref{tbl:error} we compare error types made on \beatest for T5 base and T5 xxl, trained on either \lang or \clangg. We see that for both data conditions increasing the model size leads to an increase in performance.
Comparing \clangg and \lang, shows that \clangg improves on all error types apart from orthographic (ORTH) and punctuation (PUNCT).

In Table \ref{tbl:clang8-mt5-results}, we evaluate mT5 trained on the German and Russian portions of the \clangg dataset, which contain 114K and 45K training examples, respectively. We see that for both languages performance increases with the model size, with no indication of slowing, suggesting further headroom for improvement. For German, the xxl model achieves a better score than the previous state-of-the-art, however, it is worse than \GT xxl. Whereas for Russian, mT5 trained on \clangg does not match state-of-the-art performance. We believe this is in part due to the small size of \clangg in Russian. Additionally, the training data for Russian and German comes from the same dataset as the test data which is not the case for English, making the training data of significantly greater relevance. For German and Russian GEC tasks, where in-domain training data is unavailable, \clangg could have a greater impact.

We release the re-labeled \clangg dataset, which contains 2.4M training examples for English, 114k examples for German, and 45k examples for Russian. The Czech portion of Lang-8 would have resulted in only 2k examples, and as such is excluded.

\section{Conclusion}
In this paper we report new state-of-the-art results on GEC benchmarks in four languages we studied. Our simple setup relies on a language-agnostic approach to pretrain large multi-lingual language models. To enable the distillation of our largest model into smaller, more efficient models, we released a cleaned version of the \lang dataset, enabling easier and even more accurate training of GEC models.

\section*{Acknowledgements}

We would like to thank Costanza Conforti, Shankar Kumar, Felix Stahlberg and Samer Hassan for useful discussions as well as their help with training and evaluating the models.

\bibliographystyle{acl_natbib}
\bibliography{anthology,acl2021}


\end{document}